\pdfoutput=1
\documentclass{article}

\usepackage[preprint]{neurips_2022}

\usepackage[T1]{fontenc}    
\usepackage{hyperref}       
\usepackage{url}            
\usepackage{booktabs}       
\usepackage{amsfonts}       
\usepackage{nicefrac}       
\usepackage{microtype}      
\usepackage{xcolor}         
\usepackage{booktabs}
\usepackage{makecell}
\usepackage{threeparttable}
\usepackage{amsmath}
\usepackage{graphicx}
\usepackage{wrapfig}
\usepackage{float}

\title{Dynamic Stochastic Ensemble with Adversarial Robust Lottery Ticket Subnetworks}

\author{%
    Qi Peng,\textsuperscript{\rm 1}
    Wenlin Liu,\textsuperscript{\rm 2}
    Ruoxi Qin,\textsuperscript{\rm 1}
    Libin Hou,\textsuperscript{\rm 1} 
    Bin Yan,\textsuperscript{\rm 1}
    Linyuan Wang\textsuperscript{\rm 1}
\thanks{Corresponding author.}\\
\textsuperscript{\rm 1} PLA Strategy Support Force Information Engineering University\\
\textsuperscript{\rm 2}University of Science and Technology of China (USTC)\\
  \texttt{pqmailkwiki@163.com, wanglinyuanwly@163.com } \\
}

\begin{document}

\maketitle

\begin{abstract}
Adversarial attacks are considered the intrinsic vulnerability of CNNs. Defense strategies designed for attacks have been stuck in the adversarial attack-defense arms race, reflecting the imbalance between attack and defense. Dynamic Defense Framework (DDF) recently changed the passive safety status quo based on the stochastic ensemble model. The diversity of subnetworks, an essential concern in the DDF, can be effectively evaluated by the adversarial transferability between different networks. Inspired by the poor adversarial transferability between subnetworks of scratch tickets with various remaining ratios, we propose a method to realize the dynamic stochastic ensemble defense strategy. We discover the adversarial transferable diversity between robust lottery ticket subnetworks drawn from different basic structures and sparsity. The experimental results suggest that our method achieves better robust and clean recognition accuracy by adversarial transferable diversity, which would decrease the reliability of attacks.
\end{abstract}

\section{Introduction}
Deep neural networks (DNNs) currently define state-of-the-art performance in standard image classification tasks. However, Szegedy et al. proposed that the advanced classifiers may be fooled by an imperceptible perturbation called adversarial samples\cite{szegedy2013intriguing}. It raises concern about the intrinsic vulnerability of DNNs\cite{goodfellow2014explaining,carlini2017towards}.

The defenders try hard to gain the initiative in the adversarial arms race to resist the rapid development of adversarial attacks. Researchers propose many empirical and certified defense methods to obtain robust networks. The certified defense methods are supported by rigorous theoretical security guarantees, which could steadily expand the robustness radius. However, transferring it to large datasets is not accessible due to the high computational cost\cite{tjeng2017evaluating}. And adversarial training is currently the most flexible and effective empirical defense method by enhancing the training set with adversarial samples generated dynamically\cite{madry2017towards}. Nevertheless,  previous research has testified the widespread transferability of adversarial examples\cite{papernot2016transferability,ilyas2019adversarial,inkawhich2020transferable}. And there is a demand for implicit transferability in further research because adversarial training depends on specific attack algorithms for augmented data\cite{hendrycks2021unsolved}, which makes the defender hard to and appear passive in the arms race. On the contrary, a rational ensemble strategy is an effective defense method in practice\cite{kurakin2018adversarial,liu2020enhancing}. A recent study presents the dynamic defense framework (DDF) based on the stochastic ensemble\cite{qin2021dynamic}. The DDF would change the ensemble states based on variable model attributes of the architecture and smooth parameters. It expects heterogeneous candidate models to ensure diverse ensemble statuses.


We propose the dynamic stochastic ensemble with adversarial robust lottery ticket subnetworks. Based on \emph{the Lottery Hypothesis}\cite{frankle2018lottery}, Fu et al. discover the subnetworks with inborn robustness. It matches or surpasses the adversarially trained networks with the same structure without any model training.\cite{fu2021drawing}. Inspired by Fu et al., our method obtained subnetworks with different network structures and remaining ratios to promote the adversarial transferable diversity for the DDF. By weakening the transferability between ensemble states, we improve the initiative of the DDF against the adversary.
 

\section{Method}
\label{gen_inst}
In the framework of dynamic defense, we represent the dynamic stochastic ensemble with adversarial robust lottery ticket subnetworks. \cite{fu2021drawing} proved the poor adversarial transferability between the scratch tickets under a single structure. Drawing inspiration from prior works, we further explore the adversarial transferable diversity from the different fundamental structures and remaining ratios.

\subsection{The Dynamic Defense Framework and Adversarial Transferable Diversity}
The DDF is a randomized defense strategy to protect ensemble gradient information, and the essential requirements for it are randomness and diversity to promote the ensemble's adversarial robustness\cite{qin2021dynamic}. It presents a model ensemble defense method with randomized network parameter distribution specialty, which causes an unknowable act of the defender. The output of dynamic stochastic ensemble model $f_{ens}$ containing \textit{I} number of models is defined as follows:
\begin{equation}
\label{equa1}
{f_{ens}}(x,\theta ) = \sum\limits_{i = 1}^I {f(x,\theta)} 
\end{equation}
The randomness is achieved by transferring the ensemble states with ensemble variables $\theta$. The DDF demands the construction of diversified ensemble statuses with a heterogeneous model library. Relevant studies highlight that diverse network structure plays a crucial role in ensemble defense\cite{yang2020dverge}. In our solution, we evaluated the heterogeneousness and diversity between ensemble subnetworks by the poor adversarial transferability of the attacks.

\subsection{Adversarial Robust Lottery Subnetwork}
For purpose of testifying the multi-sparsity adversarial robust lottery subnetworks can achieve better adversarial transferable diversity under different network structures. We picked four representative network structures, ResNet18, ResNet34, WideResNet32, and WideResNet38\cite{he2016deep,zagoruyko2016wide}, as the basic architecture of our experiments and gained the sparse lottery ticket from original dense networks. Following \cite{fu2021drawing}, we applied adversarial training to gain robustness of our subnetworks during pruning. It can be expressed as a min-max  problem  as Eq.\ref{equa2}.

\begin{equation}
\label{equa2}
  \mathop {\arg \min }\limits_\lambda  \sum\limits_i {\mathop {\max }\limits_{\parallel \delta \parallel  \le \varepsilon } l(f(\hat \lambda  \odot \omega ,{x_i} + \delta ),{y_i}){\rm{  }} \quad s.t.\quad  {\rm{  }}\parallel \hat \lambda {\parallel _0} \le k} 
\end{equation}

Where \textit{l} presents the loss function, \textit{f} is a randomly initialized network with random weights, and $\delta$ is the perturbation with maximum value $\varepsilon$. In order to satisfy the sparsity of the subnetworks, we set a learnable weight $\lambda$ and a binary weight $\hat \lambda  \in \{0,1\}$ that correspond to its dimensions\cite{sehwag2020hydra,ramanujan2020s}. $\hat \lambda$ is meant to activate a small number of primary weights $\omega$. With the primary network parameters weighted by $\lambda  \in  (0,1)$, \ \textit{f} can be effectively trained by small perturbations added to the input $x_i$.

\subsection{Dynamical Ensemble for The Lottery Subnetworks}\label{step}
Through our method, we obtained fourty subnetworks with different basic structures and sparsity. Based on the robust lottery subnetwork library, we define the randomized ensemble attribute parameter $\theta=\theta(\alpha,n,s)$, which determines the ensemble states. It can be achieved in the following steps:

(A)Construct a robust lottery subnetworks library with adversarial transferable diversity, including forty sparse subnetworks. Each of ResNet18/ResNet34/WideResNet32/WideResNet38 owns ten.

(B)Set the range for $\alpha$ and \textit{s}. We brought four basic structures into the selection rather than the entire library, increasing the possibility of including more structures. It realized by $\alpha=\{\alpha_i \in  \{0,1\}|i=1,...,4\}$, randomly assigned to determine the corresponding structure chosen. One means selected, and 0 means rejected. $s_k$ represents the distribution of the sparsity of each candidate structure, and each sparsity also refers to the corresponding subnetwork. Under every candidate structure, there are \textit{k} number of sparse subnetworks whose $s=\{s_k \in \{7\%, 10\%, 12\%, 15\%, 20\%, 30\%,40\%, 50\%, 60\%, 70\% \}|k=1,...,10\}$.

(C)Randomly select ensemble number $n_i$. $n_i$ presents the chosen number for each candidate structure. We set $n_i=\{n_i \in \{1,2\}|\alpha_i=1\}$ as the fraction of the total ensemble number $n$, and $n_i=0$ when $\alpha_i=0$. In particular, we gave a higher probability to smaller $n_i$, whose $p(n_i)=\{65\%,35\%|n_i \in \{1,2\}\}$. Since our experiments fed the attacker with the structure and sparsity of ensemble subnetworks, we expect to reduce the probability of a universal adversarial sample through our probabilistic solution\cite{moosavi2017universal}.

(D)Set $\theta(\alpha,n,s)$ by $n_i$ and $s_k$. According to $n_i$ determined by the distibution of $p(n_i)$, we got total ensemble number $n = \sum\limits_{i = 1}^4 {{\alpha _i}{n_i}}$. Meanwhile, randomly select $n_i$ ensemble sparsity from $s_k$, representing the corresponding subnetworks attending the ensemble.

(E)According to the ensemble variable $\theta(\alpha,n,s)$, we make the ensemble in the light of Eq.\ref{equa1}.

\section{Experiments and Results}
\label{headings}
In this section, we verify the widespread existence of robust lottery tickets and diversified adversarial transferability across different basic structures and sparsity. Then we design an adaptive attack on top of PGD-20 to evaluate the adversarial ensemble robustness of our method on CIFAR-10. 

\subsection{The Adversarial Transferability between Robust Lottery Ticket Subnetworks}
We collect forty robust lottery ticket subnetworks with different sparsity based on ResNet18, ResNet34, WideResNet32, and WideResNet38, ten of each basic structure. As shown in Tab.\ref{table1}, we marked clean accuracy and robust accuracy against PGD-20 with $\epsilon$=8 and illustrated the existence of adversarial robustness between lottery ticket subnetworks for different structures.

\begin{table}[ht]
	\caption {The clean and robust accuracy reached by the lottery ticket subnetworks library on CIFAR-10.}
	\label{table1}
\begin{tabular}{ccccccc}
    \toprule[2pt]
    Structures & Sparsity & Num   &\makecell[c]{Clean Acc\\Range(\%)} &\makecell[c]{Avg Clean\\Acc(\%)} &\makecell[c]{Robust Acc\\Range(\%)} &\makecell[c]{Avg Robust\\Acc(\%)} \\
    \midrule[2pt]
    ResNet18 &\makecell[c]{ 0.07,0.1,0.12,}     & 10    & 76.8-79.8 & 77.9  & 45.1-47.3 & 46.3 \\
    Resnet34 & \makecell[c]{0.15,0.2,0.3,}      & 10    & 77.6-80.1 & 79.1  & 46.1-48.6 & 47.6 \\
    WideResnet32 & \makecell[c]{ 0.4,0.5,}      & 10    & 79.2-82.4 & 81.1  & 48.5-49.6 & 49.1 \\
    WideResnet38 &  \makecell[c]{ 0.6,0.7}     & 10    & 79.9-83.1 & 81.9  & 49.2-50.3 & 49.7 \\
    \bottomrule[2pt]
\end{tabular}
\end{table}

\begin{wrapfigure}{tr}{0pt}

  \centering
     \includegraphics[width=2.9in,height=1.8in]{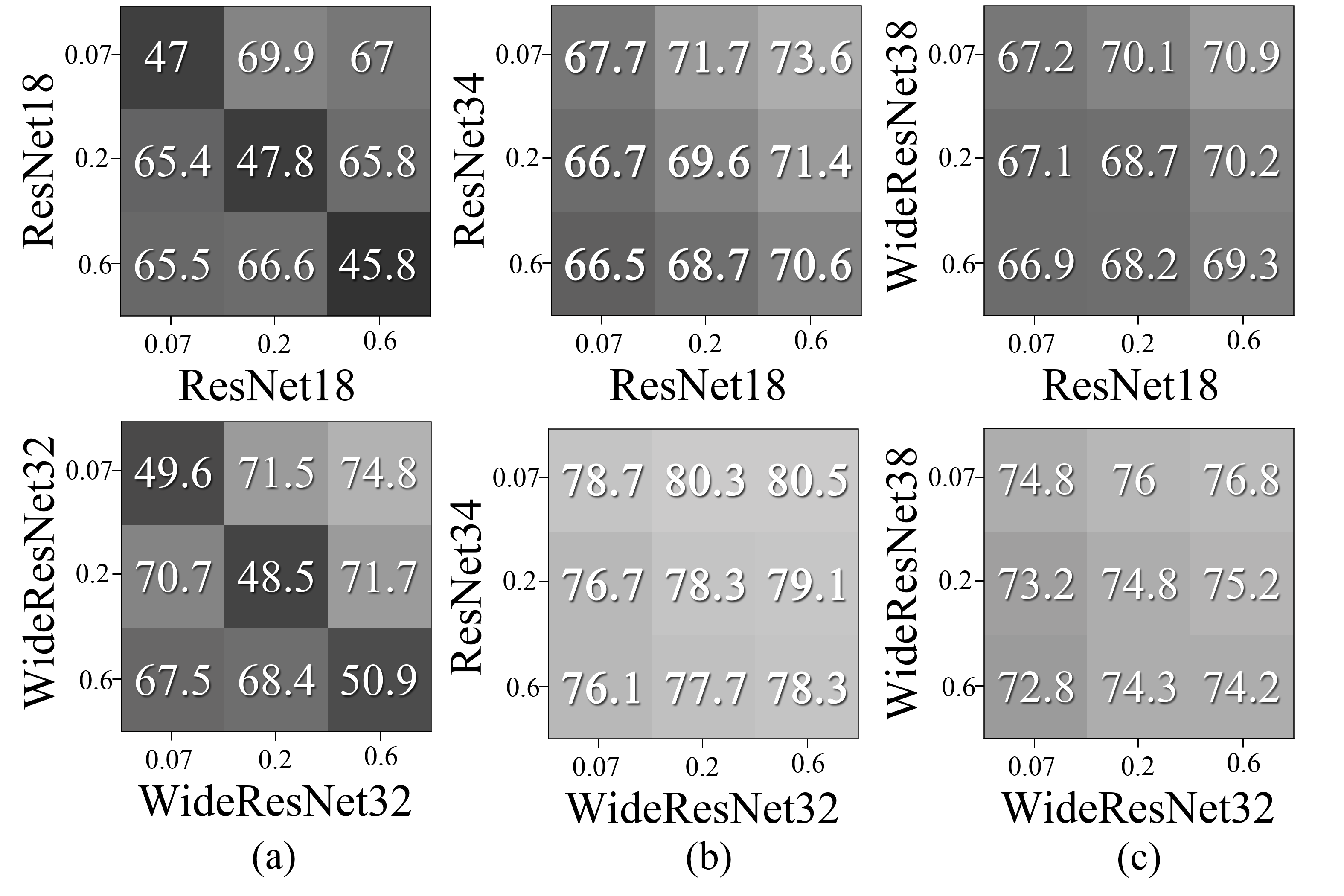}
  \caption{The adversarial transferability between subnetworks, where (a): Attack by itself; (b): Attack by ResNet34; (c): Attack by WideResNet38.}
\label{fig1}
\end{wrapfigure}
Fig.\ref{fig1} presents the pair-wise adversarial transferability with our lottery ticket subnetworks library tested under the same $\epsilon$. We adopt PGD-20 attacks with $\epsilon$=8 by constraint of $l_{inf}$. To make a fair comparison, we choose ResNet18 and WideResNet32 subnetworks with different sparsity as defense models, respectively, and pick Resnet34 and WideResNet38 subnetworks with the same sparsity to generate adversarial samples. And the distribution of sparsity with 0.07, 0.2, and 0.6 is set for models. The number represents the robust accuracy of the defense models against transferal attacks with different structures.  

As shown in Fig.\ref{fig1}, ResNet18 and WideResNet32 subnetworks with different sparsity possess poor adversarial transferability against adversarial samples generatd with the same network. Compared with (b) and (c), it is said that combination of different structures' subnetworks could weaken the adversarial transferability for attacks. E.g., except for the diagonal number, the accuracy of ResNet18 against adversarial samples generated by the same structure is 65.5\%$\sim$69.9\%. It raises to 66.5\%$\sim$73.6\% and 66.9\%$\sim$70.9\% facing transferable attacks by ResNet34 and WideResNet38 subnetworks. Likewise, WideResNet32 subnetworks' accuracy was raised for 5.7\%$\sim$9.3\% and 2\%$\sim$5.9\%.

\subsection{Emsemble Robustness}
\begin{wrapfigure}{tr}{0pt}
\centering
\includegraphics[width=0.4\textwidth]{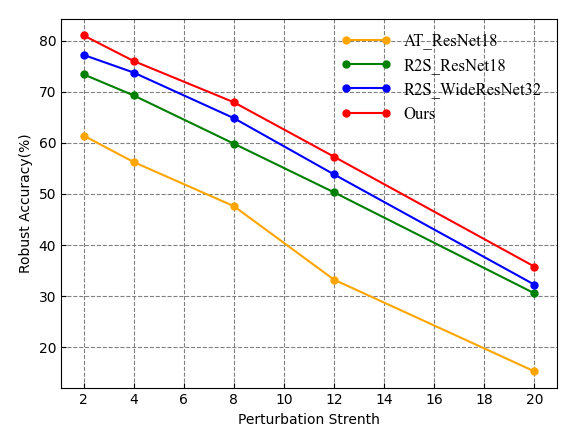}   
\caption{\footnotesize Comparing the robustness of our method with adversarial training and R2S against the EOT attacks on CIFAR-10.}
\label{fig2}
\end{wrapfigure}
In this section, we validate the effectiveness of our defense strategy. We set the adversarial training as the baselines and compare our method with R2S\cite{fu2021drawing}, which ensemble different remaining ratios from the same networks. 

\textbf{Evaluation setup}. Since both the R2S and our method could adjust the probability for their sparsity choices, we assume that both adopt uniform sampling from the same sparsity with \ref{step} for simplicity. Moreover, we design an adaptive attack based on  the \emph{Expectation over Transformation(EOT)}\cite{athalye2018synthesizing} that generates adversarial examples via the expectations of the gradients from all candidate robust lottery subnetworks, which achieves the attack effect by promoting the transferability of adversarial samples, traversing the possibility of defense strategy.





We set adversarial trained ResNet18/WideResNet32 dense networks as baselines and compared our method with the R2S. In order to comprehensively and accurately observe the defense effect, we adopt multiple $\epsilon$=[ 0, 2, 4, 8, 12, 20] with white-box attack in the $l_{inf}$ norm. For the EOT attack, we announced the network structures and remaining ratio of the lottery ticket library so that the attacker could sample the expectation of different ensemble statuses.


\begin{table}[ht]
    \centering
	\caption {Comparing the robustness of our method with adversarial training and R2S against the EOT attacks on CIFAR-10. Our method ensemble four basic structures.}\begin{tabular}{ccccc}
    \toprule
    Network & Resnet18 & WideResnet32 & Resnet18 & WideResnet32 \\
    \midrule
    Method & clean acc(\%) & clean acc(\%) & robust acc(\%) & robust acc(\%) \\
    \midrule
    Dense & 81.73 & 85.93 & 51.2  & 52.3 \\
    R2S   & 78.06 & 82.34 & 57.6  & 64.7 \\
    Ours  & \multicolumn{2}{c}{87.01} & \multicolumn{2}{c}{67.72} \\
    \bottomrule
    \end{tabular}%
  \label{tab2}%
\end{table}

As shown in Tab.\ref{tab2}, the robust accuracy of our method is 3.02\%$\sim$10.12\% higher than R2S. Meanwhile, R2S dropped 3.59\%$\sim$3.67\% in clean accuracy, while the dynamic ensemble for different structures raised 1.08\% compared to adversarial training. For adversarial training, our method got a 15.42\% raise over the robust accuracy. In addition, Fig.\ref{fig2} shows that our method has better adversarial robustness than the R2S and adversarial training under the overall environment with multiple perturbations.

\section{Conclusion}
\label{others}
In this paper, we propose the dynamic ensemble method based on adversarial lottery ticket subnetworks, which describes how the diversity of ensemble robustness is presented as adversarial transferability among subnetworks. We gather different basic structures and sparsity for each robust lottery ticket subnetwork. Furthermore, we make poor adversarial transferability and diversified ensemble statuses between models by picking stochastic ensemble models. Our experiments show that diversified structures and sparsity of scratch tickets weaken the adversarial transferability for subnetworks and improve the adversarial robustness of the ensemble method.

\bibliographystyle{unsrt}
\bibliography{neurips_2022}

\end{document}